# nD-PDPA: nDimensional Probability Density Profile Analysis

**Arjang Fahim**[1], **Stephanie Irausquin**[1], **Homayoun Valafar**[1]

[1]Department of Computer Science and Engineering, University of South Carolina, Columbia, SC, USA

*Author to whom correspondence should be addressed; E-Mail: homayoun@cec.sc.edu;
Tel.:+1-803-777-2404;Fax:+1-803-777-3637

## 1- Abstract

Despite the recent advances in various Structural Genomics Projects, a large gap remains between the number of sequenced and structurally characterized proteins. Some reasons for this discrepancy include technical difficulties, labor and the cost related to determine a structure by experimental methods such as NMR spectroscopy. Several computational methods have been developed to expand the applicability of NMR spectroscopy by addressing temporal and economical problems more efficiently. While these methods demonstrate successful outcomes to solve more challenging and structurally novel proteins, the cost has not been reduced significantly.

Probability Density Profile Analysis (PDPA) has been previously introduced by our lab to directly address the economics of structure determination of routine proteins and identification of novel structures from a minimal set of unassigned NMR data. 2D-PDPA (in which 2D denotes incorporation of data from two alignment media) has been successful in identifying the structural homologue of an unknown protein within a library of ~1000 decoy structures. In order to further expand the selectivity and sensitivity of PDPA, incorporation of additional data was necessary. However, expansion of the original PDPA approach was limited by its computational requirements where inclusion of additional data would render it computationally intractable. Here we present the most recent developments of PDPA method (nD-PDPA: n Dimensional Probability Density Profile Analysis) that eliminate 2D-PDPA's computational limitations, and allow inclusion of RDC data from multiple vector types in multiple alignment media.

## 2- Introduction

The ultimate goal of Structural Genomic Projects are to identify protein structures, their function and biological significance for newly discovered sequences [1]. To achieve this goal, development of accurate methods to correctly predict the three-dimensional structure of an unknown protein is crucial. Currently existing protein structure prediction methods can be categorized into two broad major groups: homology and ab-initio modelings.

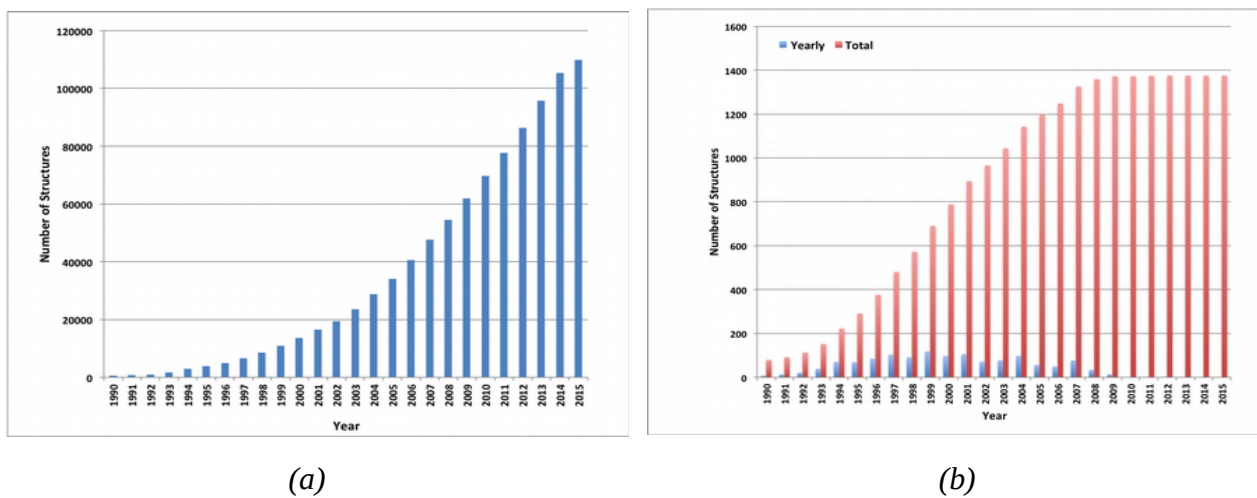

*(a)*  *(b)*

**Figure1:** *A number of protein structures in PDB (a) cumulative since 1990 and (b) unique folds reported by SCOP.*

Homology modeling is based on selecting one or more frame (template) structures from protein databases that resemble the query structure's sequence [2]. The success of any template based method is heavily dependent on completion of the protein structure databases. Statistics from Protein Data-Bank [3], [4](http://rcsb.org) indicates the total number of 109,093 (May 2015) protein structures deposited to the database. In Figure1(a), it is clear that the number of novel proteins deposited has been increasing over the past decade, however no new fold-family has been characterized since 2010 (Figure1(b)) and the number of fold families remain constant at only about 2500 despite being previously predicted as 10,000 protein fold families [5]. This indicates that traditional homology methods for identification of a novel protein structure based on sequence homology, will inherently fail due to the lack of sufficient structural information.

If no such template exists for a desired unknown structure within a protein library, then three-dimensional structure of the unknown protein may potentially be calculated from ab-initio modeling [6]–[8] tools.

In this manuscript, we present n-Dimensional Probability Density Profile Analysis (nD-PDPA), an enhanced version of the previously reported 2D-PDPA for identifying and characterizing a protein structure that utilizes unassigned Residual Dipolar Coupling (RDC) data [9], [10]. RDCs are relatively rapid to obtain and they contain orientational information about internuclear vectors relative to an arbitrarily selected molecular frame [11], [12]. Utilization of large variety of alignment media (such as PEG and lipid bicells)[13], [14] increases the practical usage of RDC data as a stand alone source of data as well as a complement source of information to other NMR data sources such as NOE and J-Coupling in refinement of structures [15] .

Recent improvements in computational modeling introduces more advanced techniques that combines different modeling methods with experimental constraints data. CS-ROSETTA [16] and RDC-ROSETTA [17] are two examples of such approaches. Although these methods may alleviate the problem of selecting the target template structures, they rely on assigned experimental data that is labor intensive and costly [18], [19]. In PDPA however, unassigned sets of RDCs are utilized as main source of data. The utilization of unassigned RDC sets eliminates the time-consuming assignment process. The application of 2D-PDPA method utilizing two sets of RDCs has been demonstrated[10]. However, 2D-PDPA method is limited to two homogeneous sets of RDC data, usually N-H vectors from two different alignment media. nD-PDPA, on the other hand is capable of utilizing more than two sets of unassigned RDC data from different alignment media. Moreover, the type of RDC vectors used in nD-PDPA are not limited to N-H vectors and can be the combination of different vectors such as N-H and Cα-Hα. Addition of more than two sets of RDCs from different vector types greatly increases the accuracy and selectivity of the method. However, this transition to higher dimensional data introduces new challenges such as intractable program execution time. Therefore, the core engine of 2D-PDPA method has been revised to fulfill new requirements. The nD-PDPA method has been developed in C++ utilizing Object Oriented paradigm and can be executed on desktop machines or Linux clusters. Execution on Linux clusters is facilitated by the use of qSub protocol. The source code for nD-PDPA can be obtained from our website at http://ifestos.cse.sc.edu.

In this report, the PDPA method is reviewed briefly then the result of 2D-PDPA and nD-PDPA are compared for accuracy, sensitivity and measure of execution time. Finally, the nD-PDPA method is validated by utilizing protein structures varying in size and secondary structures with synthetic data.

## 3- Residual Dipolar Couplings

The phenomena of Residual Dipolar Couplings was established for the first time in 1960's [20]. However the practical application of RDCs to Biomolecular systems emerged a decade ago for the first time [21], [22] . Since then, many methods have been developed to utilize RDC data in wide range of automated backbone resonance assignments, structure determination, protein folding to ligand protein, protein-protein interaction and protein dynamics [23], [24]. In this manuscript, we will not intend to review the practical and instrumental aspect of the RDC acquisition. Instead, we concentrate on the mathematical interpretation of the RDC for our study purpose. Interested reader can be directed to [25], [26] for more information.

The physical basis of Residual Dipolar Couplings is the dipole-dipole interaction between two nuclear spins. In the presence of an external magnetic field the RDC between two spins i and j is given by Equation (1):

$$D_{ij} = \frac{-\mu_0 \gamma_i \gamma_j h}{8\pi^3 r^3} \left\langle \frac{3\cos^2\theta - 1}{2} \right\rangle \qquad (1)$$

where $\gamma_i$ and $\gamma_j$ are magnetogyric ratio of given nuclei, $h$ is Plank's constant, $r$ is the distance of two nuclei, and $\theta$ is the angle between internuclear vector and external magnet field $B_0$.
In an isotropic tumbling system the RDC values averages to zero and no RDC is observed. This can be treated by introducing alignment medium to partially align the system for the observable RDC values.
Equation (1) can be reformulate by algebraic manipulation suitable for computational purpose as below:

$$D_{ij} = D_{max} v_{ij} R(\alpha, \beta, \gamma) \begin{pmatrix} S_{xx} & 0 & 0 \\ 0 & S_{yy} & 0 \\ 0 & 0 & S_{zz} \end{pmatrix} R(\alpha, \beta, \gamma)^T v_{ij}^T \qquad (2)$$

where $v_{ij}$ is the inter-nuclear vector between atoms $i$ and $j$. $R(a, b, g)$ describes an Euler rotation of the molecule with respect to Principle Alignment Frame(PAF). The parameters $S_{xx}$, $S_{yy}$, and $S_{zz}$ are known as Principle Order Parameters (POP) that describe the strength of alignment along each of the axes of alignment.

**4- Methods**
nD-PDPA is an extension of the previously reported 2D-PDPA [10] that allows simultaneous analysis of more than two sets of RDC data. In principle, PDPA does not have to be limited to two sets (as in 2D-PDPA) or homogeneous data types (e.g. using only Cα−Hα RDC sets). Normally multiple RDC data types such as N-H or Cα−Hα, can be acquired in one experimental session therefore inclusion of which impose no additional data acquisition time. Integration of additional data is expected to substantially increase the information content and therefore significantly improve the sensitivity and robustness of the PDPA method. The details of the PDPA method are described previously [10], [27], [28]. In this manuscript we provide a brief overview of the PDPA method and focus primarily on the new additions and improvements of the nD-PDPA. The core principle of the PDPA method is based on the fact that two similar structures should produce the same distribution patterns of RDCs, and can be used as structural finger print. Therefore, measurement of the similarity between two distributions can be interpreted as similarity of two structures.
The nD-PDPA algorithm is encapsulated in three functional layers. In the first layer, the experimental RDC sets are used to estimate the relative order tensors [29], [30]. The number of parameters needed in this stage is a function of the number of alignment media in which RDC data are acquired. Generally for RDC data from n alignment media, 5n-3 parameters are required to describe the relative order tensors [31]. The estimated order tensor parameters are utilized to back calculate the RDC data for a given structure. Then the n-dimensional Kernel Density Estimation is utilized to construct the distribution map for both experimental and calculated RDC sets. The kernel Density Distribution is calculated based on a hyper dimensional Gaussian Kernel function (described in Equation 3) that is located at the center of each RDC data point (Figure2).
In this equation X denotes independent function parameters and M denotes vector of RDCs that defines the center of the kernel and Σ is covariance matrix. Both X and M vectors are of size k while the Σ is a matrix of size k x k:

$$N(X \vee M, \Sigma) = (2\pi)^{-k} \|\Sigma\|^{-k} \exp[-1/2(X-M)'\Sigma^{-1}(X-M)] \qquad (3)$$

The orientation of the anchor alignment medium [10], [32],[31] is exhaustively searched with respect to the reference structure. Therefore in the second stage the PDP map is calculated for the subject structure in every possible orientation using a grid search over the Euler angles (*a, b, g*) at the resolution of 5˚. The best score is calculated from comparison of the experimental and calculated RDC distributions for all orientations and this score is reported as the final result in the third stage. For a given library of the structures the process is repeated for every structures in the library.

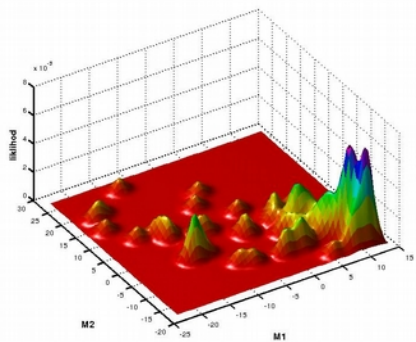

*Figure2:* An example of a 2D-PDP map, using kernel density estimation. This 2D-PDP can serve as a structural finger print.

## 5- Scoring of nD-PDPA

Manhattan (City Block) metric [33] (shown in Equation 4) had been utilized in 2D-PDPA to compare calculated and experimental probability density distributions. In this equation $B_{score}$ denotes the 2D-PDPA score; the summation indexes i and j cover the entire range of RDCs over alignment media M1 and M2; and $cPDP_{ij}$ (calculated PDP) and $ePDP_{ij}$ (experimental PDP) denote the likelihood of the RDC values at the location i and j. 2D-PDPA utilizes the locations of i,j that are represented by a 64 × 64 grid. This grid is constructed by uniformly sampling the entire range of both RDC sets for both ePDP and cPDP. Utilization of the grid guarantees the similar intervals and range (begins with minimum RDC and ends with maximum RDC values) for both cPDP and eDPD.

$$B_{Score} = \sum_{Min(M2)}^{Max(M2)} \sum_{Min(M1)}^{Max(M1)} |cPDP_{ij} - ePDP_{ij}| \quad (4)$$

In order to be qualified for probability density functions, the summation of ePDP and cPDP for the entire range of RDC values are normalized to be zero. Therefore the $B_{score}$ ranges from [0-2]. The $B_{score}$ of 2 refers to completely dissimilar structures and $B_{score}$ of 0 refers to 100% structural similarity. The other factors such as RDC error and availability of data also effects in the $B_{score}$.

Comparison of ePDP and cPDP in a grid fashion is the main contributing factor for the exponential time-complexity of this approach. In that sense, expansion of ePDP and cPDP patterns to n-dimensions requires an exponentially increasing number of grid points ($64^n$ if 64 points along each dimension) to serve as the location of comparisons by factor of grid size. This quickly becomes a limiting factor for n > 2. Moreover, since RDC data are not uniformly distributed [34], any PDP distribution will contain large areas with likelihood of zero or near zero. Incision of these unimportant regions for comparison of two PDPAs consume unnecessary computational time (Figure3). The score in nD-PDPA is on the other hand, calculated by comparison of only the information rich regions within the distributions. By using this approach the regions with likelihood of zero or close to zero are not considered for calculation and therefore exponential contribution of the grid size is eliminated.

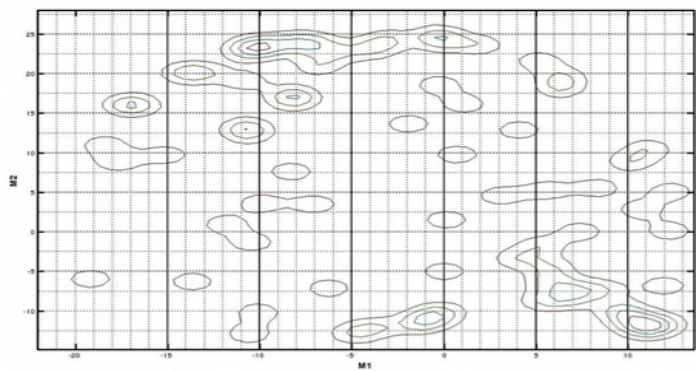

*Figure3:* 2D-PDPA utilizes a 64 by 64 grid for both computational and experimental RDC sets for scoring. The out of boundaries area are unnecessary for calculation.

## 6- Data Preparation

The three structures shown in Table 1 are used throughout our experiments. These proteins have been selected on the basis of secondary structure composition that represent the three broad structural categories of α, β and α/β (Figure4). The atomic coordinates of the structures have been obtained from the Protein Data-Bank [3]. Three sets of RDCs including N-H vectors representing first alignment medium and N-H and Cα-Hα vectors representing second alignment medium are generated under two conditions: 1- Ideal RDC sets containing no error, 2- corrupted RDC sets through the addition of ±1 Hz of uniformly distributed noise with 25% of RDCs randomly eliminated from each set to better represent pragmatic conditions. The first set (no error) that represents the ideal conditions is utilized for demonstrating the proof of concept and the second set represents a more realistic conditions. To generate synthetic RDC sets the software REDCAT [33] was used with the initial relative order tensors listed in Table 2.

*Table 1:* Protein structures obtained from Protein Data Bank.

| Protein | Secondary Structure | Number of Residues | CATH Classification |
|---|---|---|---|
| 1A1Z | α | 83 | 1.10.553.10 |
| 1OUR | β | 114 | 2.60.120.40 |
| 1G1B | α/β | 164 | 3.40.1410.10 |

Upon completion of the data generation, the assignment information is discarded before utilization of the synthetic RDC data in nD-PDPA. To back calculate the order tensors two approaches were used: First the optimal order tensor is calculated using REDCAT and second estimation of the order tensors were conducted using approximation method as described previously [35].

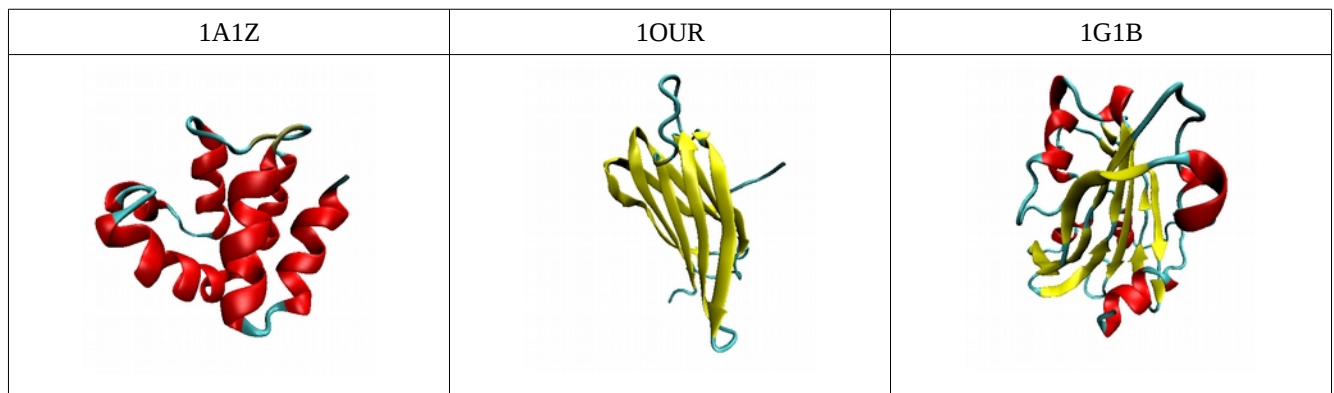

*Figure4 :* Cartoon representation of the proteins used in experiment.

*Table 2: List of Initial order parameters generated by REDCAT, used to calculate RDC sets.*

|    | Sxx   | Syy   | Szz    | α  | β  | γ   |
|----|-------|-------|--------|----|----|-----|
| **M1** | 3e-4  | 5e-4  | -8e-4  | 0  | 0  | 0   |
| **M2** | -4e-4 | -6e-4 | 1.00e-4 | 40 | 50 | -60 |

Estimation of the order tensor parameters is of central importance for PDPA analysis in the absence of atomic coordinates of a structure. 2D and 3D approximation software were used [30] to estimate relative order tensor that can be employed in PDPA analysis. In nD-PDPA experiments relative order tensors are estimated utilizing approximation when the data are synthetically corrupted by adding errors (Table 3).

*Table 3: Order Tensor parameters estimation using 2D-Approax software for the data in Table 2. The data is corrupted by ±1Hz of error and 25% of the RDCs are removed from datasets.*

|    | Sxx         | Sxy           | Sxz          | Syy          | Syz           |
|----|-------------|---------------|--------------|--------------|---------------|
| **M1** | 0.00028136  | -5.38223e-07  | 4.70822e-07  | 0.000446853  | 1.86497e-07   |
| **M2** | 0.000428708 | -0.000417908  | 0.00022974   | 0.000726639  | -0.000129934  |

## 7- Results and Discussion
### 7.1 Experiment 1

The objective of this experiment is to establish the relation between nD-PDPA score and bb-rmsd. To accomplish this objective 1000 decoy structures were generated from the native structure by randomly altering the φ and ψ angles to generate structures with bb-rmsd in the range of 0-8Å of the native structure. The entire ensemble of the decoy structures were then subjected to evaluation by nD-PDPA. Finally, the scatter plot of bb-rmsd versus nD-PDPA scores were used to observe any significant patterns. Previously the universal funneling effect of such an exercise had been demonstrated [10] for proteins regardless of their structural characteristics. In Experiment 1 we repeat the previous exercise using nD-PDPA engine and compare some of the results with the previous PDPA program (2D-PDPA). In this experiment, the protein 1A1Z was selected with RDC data generated in REDCAT[36] with no added error. Figure5(a) shows the relationship between 2D-PDPA score and bb-rmsd for one thousand decoy structures generated from reference structure 1A1Z. The same experiment was conducted using nD-PDPA engine in Figure5(b). As it is mentioned in section 5 in nD-PDPA engine the comparison for calculated and experimental PDP is based on the RDC points and not by 64 by 64 grid as it is performed in 2D-PDPA. This may reduce the sensitivity of the scoring in nD-PDPA compared to the 2D-PDPA ( the $R^2$ fitness for 2D-PDPA is slightly better than $R^2$ in nD-PDPA analysis in Figure5). The lack of sensitivity can be addressed by adding more RDC sets improving the information content of the nD-PDPA analysis and nD-PDPA score fitness.

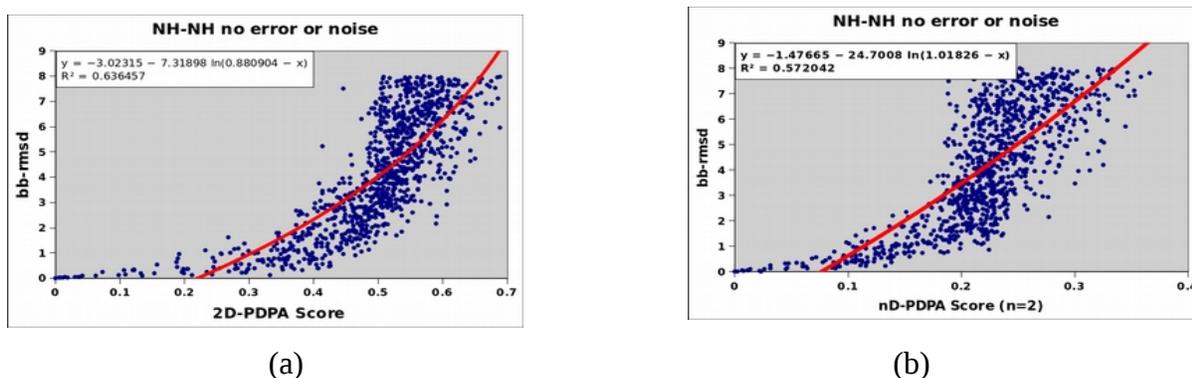

(a)                      (b)

*Figure5 :* (a) The funneling pattern of 2D-PDPA Score and bb-rmsd of 1000 decoy structures for protein 1A1Z (83) (b) The same experiment using the similar set of data with the nD-PDPA engine.

Figure6 shows the nD-PDPA analysis using three N-H RDC sets from three alignment media. The $R^2$ shows improvement, compared to both 2D-PDPA and nD-PDPA analysis using two RDC sets (Figure5(a) and (b)).

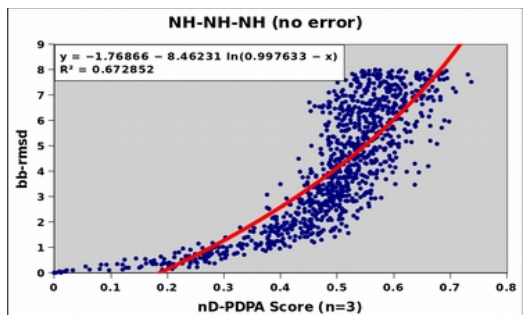

*Figure6:* The funneling pattern of nD-PDPA score and bb-rmsd of 1000 decoy structures for protein 1A1Z(83). Three N-H RDC sets were used to conduct this experiment.

In the previous experiment the improvement of $R^2$ in ideal conditions by adding more RDC data was demonstrated. It is useful to utilize RDC data that is closer to the experimental conditions. The Previous experiment was repeated by randomly removing 25% of the RDC values for protein 1A1Z. In Figure7(a) and (b), the results of 2D-PDPA and nD-PDPA (using three sets of N-H RDC vectors) analysis are demonstrated. The score and bb-rmsd fitness score shows better correlation in the case of nD-PDPA (n=3) analysis.

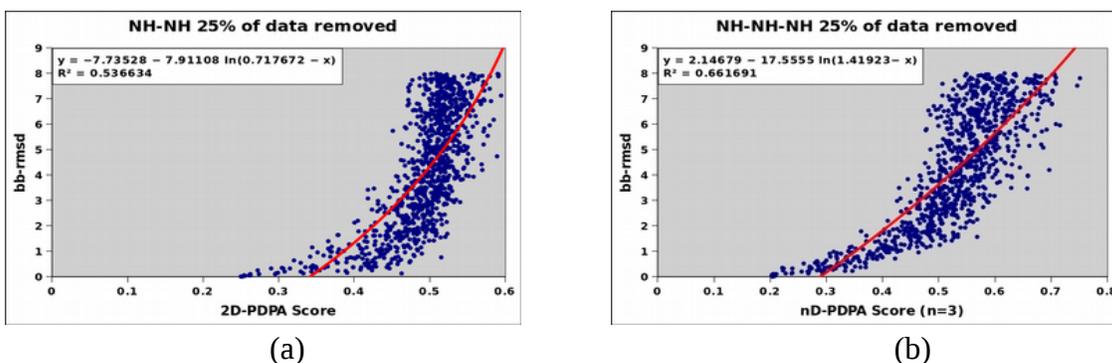

(a)            (b)

*Figure7*: 250 decoy structures (1A1Z as reference) with (a) two N-H RDC sets (2D-PDPA) (b) three N-H RDC sets (nD-PDPA). 25% of the data were randomly removed from each set.

### 7.2 Experiment 2

In Experiment 2, protein 1G1B(164) was used, and RDC datasets were corrupted by addition of ±1Hz of uniformly distributed error and randomly removing of 25% of the RDC values from each set. 250 structures were generated by altering backbone torsion angles range from 0-6Å from 1G1B.

In Figure8(a) nD-PDPA analysis was conducted by using two N-H RDC sets from two alignment media and in Figure8(b) the analysis was conducted by utilizing two non-homogeneous RDC sets, N-H and Cα-Hα from two alignment media respectively. $R^2$ values for both experiments are approximately similar (~0.7).

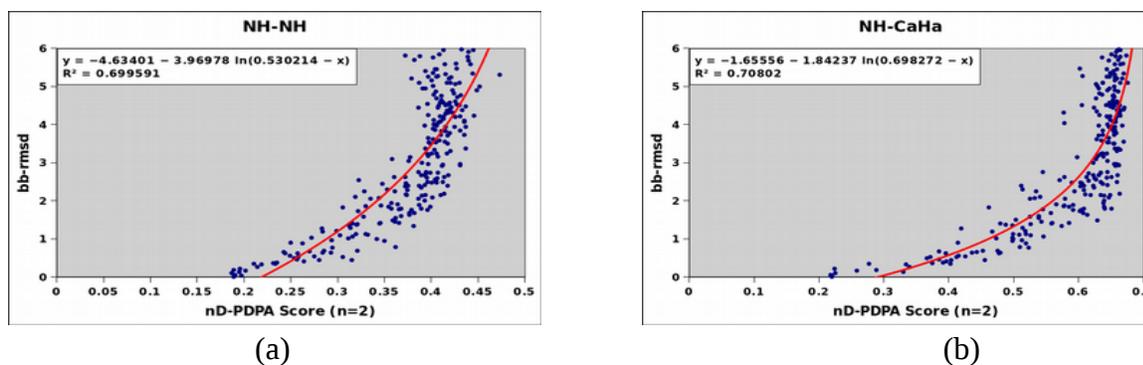

    (a)                                                       (b)

*Figure8:* (a) Calculated nD-PDPA scores vs. bb-rmsd for protein 1G1B using two N-H RDC sets. (b) Calculated nD-PDPA scores vs. bb-rmsd for protein 1G1B using two N-H and Cα-Hα RDC sets.

Figure9 shows the plot of nD-PDPA analysis for protein 1G1B by adding Cα-Hα as the third RDC set to the collection of two N-H RDC sets. The $R^2 = 0.8146$ indicates the improvement of nD-PDPA analysis fitness compared to two sets of RDCs (Figure8(a) and (b)). The experiment once again confirms the improvement of the bb-rmsd and PDPA score fitness by adding more RDC data.

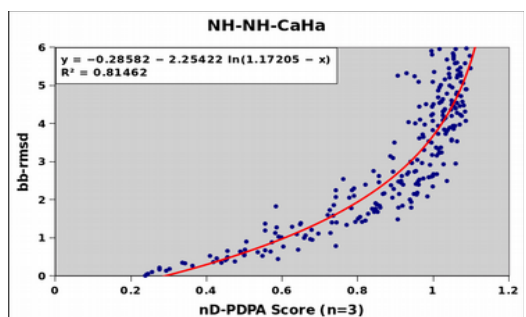

*Figure9:* Calculated nD-PDPA scores vs. bb-rmsd for protein 1G1B utilizing three sets of RDC. Two of which are N-H sets and the third one is Cα-Hα. $R^2$ value is improved in comparison to utilization of two RDC sets.

### 7.3 Experiment 3

Table 5 demonstrates the $R^2$ value for two of the proteins listed in Table 1. In Table 5 in all instances the value of the $R^2$ decreases by introducing error to the data ( compare second and third column of Table 5(a),(b)). Moreover, the value of $R^2$ increases by adding an extra set of RDC. The improvement is manifested in the greater degree in the erroneous data sets. For example in Table 5(a) $R^2$ for three N-H RDC sets demonstrates about 0.35 improvement, compared to two N-H RDC sets (See the darker background cells in Table 5(a)).

*Table 5*: The result of nD-PDPA analysis for three structures listed in Error: Reference source not found.

| Protein 1OUR | $R^2$ no error | $R^2$ with error | Protein 1G1B | $R^2$ no error | $R^2$ with error |
|---|---|---|---|---|---|
| NH-NH | 0.5781 | 0.4522 | NH-NH | 0.6924 | 0.6995 |
| NH-CαHα | 0.6797 | 0.4935 | NH-CαHα | 0.7204 | 0.7080 |
| NH-NH-NH | 0.7273 | 0.7026 | NH-NH-NH | 0.7494 | 0.7575 |
| NH-NH-CαHα | 0.6873 | 0.6547 | NH-NH-CαHα | 0.8225 | 0.8146 |

                    (a)                                             (b)

### 7.4 2D-PDPA vs. nD-PDPA running time

To benchmark the performance of the nD-PDPA for two or more sets of RDC, protein 1A1Z was selected, and the results were compared to 2D-PDPA results. Four synthetic N-H RDC sets were generated in an ideal condition. Both programs were executed on a Linux desktop with Intel Core i7, 2.67 GHz processor and 8 MB of memory. Table 6 shows the results of execution time for 2D-PDPA and nD-PDPA. Although not proven here, the asymptotic execution time of 2D-PDPA is a function of $O(C^n)$ while the execution time complexity of nD-PDPA is a function of $O(Cn^2)$. The 2D-PDPA is incapable of incorporating more than two RDC sets hence the 2D-PDPA running times for the dimensions n > 2 were approximated using the 2D-PDPA asymptotic function. For two RDC sets the running time was measured by executing the 2D-PDPA software. The running time for one RDC set was collected from 1D-PDPA version of the software and was assumed both 2D-PDPA and nD-PDPA consume the same execution time. The results indicate tremendous time reduction in nD-PDPA engine especially for n>=3.

*Table 6:* Execution time needed by 2D-PDPA and nD-PDPA.

| # of available RDC sets | ND-DPAP required time (seconds) | 2D-PDPA required time (seconds) |
|---|---|---|
| 1 | 20 | 20 |
| 2 | 323 | 363 |
| 3 | 484 | 6859 |
| 4 | 906 | 130321 |

It is worthy to note that the listed running times are only for one structure. Usually a PDPA experiment utilizes a library of structures that is indeed impossible to be finished in the reasonable time in the case of 2D-PDPA method.

### 8- Conclusion

Based on results from Experiments 1-3 a transition from 2D-PDPA to nD-PDPA can be deemed advantageous for a number of reasons. First, based on availability, additional RDC datasets can be combined from multiple alignment media to increase the information content without imposing a substantial penalty in the execution time. Second, nD-PDPA's scope of RDC analysis no longer limited to just N-H RDC data. The new improvements enable flexible inclusion of RDC data from the same or different alignment media. For example {N-H, Ca-Ha} data from one alignment medium may be combined with {N-H} of the second and {C-N} of the third alignment medium for a total of four dimensional analysis of PDPA. This flexible inclusion of any available datasets from any number of alignment media can increased the information content significantly leading to a more improved sensitivity and selectivity performance of nD-PDPA. Elimination of the exponential time-complexity and translation of the algorithm into a polynomial time-complexity is a major achievement with clear consequences in the execution time of the algorithm.

This research will be expanded in the future by utilization of experimental RDC data. It is easy to envision the application of the nD-PDPA method to characterize an unknown protein structure among a library of proteins from data-bank, only based on different types of unassigned RDC sets.